\pdfoutput=1
\documentclass[11pt]{article}

\usepackage[]{acl}

\usepackage{times}
\usepackage{latexsym}
\usepackage{microtype}
\usepackage{graphicx} 
\usepackage{xspace}
\usepackage{amsfonts}
\usepackage{subfigure}
\usepackage{amsmath}
\usepackage{mathtools}
\usepackage{arydshln}

\newcommand{\ourmodel}{Ered}

\newcommand{\openentity}{Open Entity}
\newcommand{\figer}{FIGER}
\newcommand{\fewrel}{FewRel}
\newcommand{\tacred}{TACRED}

\newcommand{\sst}{SST}
\newcommand{\eem}{EEM}

\usepackage[T1]{fontenc}
\usepackage[utf8]{inputenc}

\usepackage{microtype}

\title{\ourmodel: Enhanced Text Representations with Entities and Descriptions}

\author{Qinghua Zhao$^1$, Shuai Ma$^1$, Yuxuan Lei$^2$ \\
        $^1$SKLSDE Lab, Beihang University, Beijing, China\\
        $^2$University of Science and Technology of China, Hefei, China\\
        \{zhaoqh, mashuai\}@buaa.edu.cn\\
        lyx180812@mail.ustc.edu.cn} 
\begin{document}
\maketitle
\begin{abstract}
% 1. 我们做什么， 2. 已有方法通常做法， 3. 已有方法不足， 4. 我们的方法， 5. 实验表明
External knowledge, e.g., entities and entity descriptions, can help humans understand texts. Many works have been explored to include external knowledge in the pre-trained models. These methods, generally, design pre-training tasks and implicitly introduce knowledge by updating model weights, alternatively, use it straightforwardly together with the original text. Though effective, there are some limitations. On the one hand, it is implicit and only model weights are paid attention to, the pre-trained entity embeddings are ignored. On the other hand, entity descriptions may be lengthy, and inputting into the model together with the original text may distract the model's attention. This paper aims to explicitly include both entities and entity descriptions in the fine-tuning stage. First, the pre-trained entity embeddings are fused with the original text representation and updated by the backbone model layer by layer. Second, descriptions are represented by the knowledge module outside the backbone model, and each knowledge layer is selectively connected to one backbone layer for fusing. Third, two knowledge-related auxiliary tasks, i.e., entity/description enhancement and entity enhancement/pollution task, are designed to smooth the semantic gaps among evolved representations. We conducted experiments on four knowledge-oriented tasks and two common tasks, and the results achieved a new state-of-the-art on several datasets. Besides, we conduct an ablation study to show that each module in our method is necessary. 

\end{abstract}

%%%%%%%%%%%%%%%%%%%%%%%%%%%%%%%%%%%%%%%%%%%%%%%%%%%%
\begin{table}[t]
\centering
\resizebox{\columnwidth}{!}{
\begin{tabular}{l|l}
\hline 
Text&    The British \underline{Information Commissioner 's} \\
&   \underline{Office} invites Web users to locate its \\
&   address using Google Maps . \\ 
Mention  &    Information Commissioner 's Office\\
Span &    (12, 46)\\
Entity&  Information Commissioner's Office \\
Description&    British data protection authority\\
\hline
\end{tabular}
}
\caption{An example of a text and its associated entities and descriptions, extracted from  \openentity~dataset.}
\label{exapmle} 
\end{table}
%%%%%%%%%%%%%%%%%%%%%%%%%%%%%%%%%%%%%%%%%%%%%%%%%%%%

\section{Introduction}
% 1. PTM is popular, but ...disadvantage
Pre-trained language models (PLMs), including BERT~\citep{devlin2019bert} and RoBERTa~\citep{liu2019roberta}, have achieved state-of-the-art (SOTA) performances on various natural language processing (NLP) tasks. These PLMs can learn rich linguistic knowledge from unlabeled text~\citep{NelsonFLiu2019LinguisticKA}. However, they capture some kinds of statistical co-occurrence and cannot sufficiently capture fact or commonsense knowledge~\citep{petroni2019language, lietard2021language}. They always have better representation on popular token instead of tail token~\citep{LaurelOrr2020BootlegCT}.

% 2. some works, though effective, but ...disadvantage 
Entities and its associated descriptions in knowledge graphs (KGs), e.g., ConceptNet~\citep{speer2017conceptnet}, WordNet~\citep{miller1995wordnet}, Wikidata~\citep{vrandevcic2014wikidata} and DBpedia~\cite{brummer2016dbpedia}, just to name a few, contain extensive information. 
Table~\ref{exapmle} shows an example of a given text and its associated entities and entity descriptions (only one is shown in the table), obviously, the description can help understand.
Some works have focused on incorporating entities or entity descriptions into PLMs~\citep{xiong2019pretrained, peters2019knowledge, levine2020sensebert, zhao2022kesa, zhang2019ernie, yamada2020luke, wang2021kepler, xu2021fusing, wang2021k, xu2021does}.
Usually, they design knowledge-related pre-training tasks, e.g., entity prediction and entity relation prediction tasks, to continue pre-training the models on a large-scale corpus. External knowledge is therefore implicitly introduced by updating the models' parameters. Alternatively, they directly append the text of entities or descriptions to the original input text, treating entities or descriptions as additional text to enrich the original entry.
% disadvantage: 1. 隐式，2. entity没使用，3. des太长
Although these methods have yielded promising results, we argue that they have the following shortcomings.
Firstly, when entities and descriptions are involved in continuing pre-training, the knowledge is only implicitly injected by updating the model parameters. 
Moreover, during this process, the entity embeddings pre-trained by these pre-training tasks, which cost many computation resources and are of great value, are wasted. 
Secondly, when the entities and descriptions are appended directly to the original text, it will leads to huge costs of computing resources and a diversion of the model's attention, as  the descriptions are always long texts.

%   3. To alleviate the above issues, .
To alleviate the above issues, we propose \ourmodel, where both entities and entity descriptions are explicitly included to enhance the representation of the original text.
Firstly, the pre-trained entity embeddings are explicitly fused with the original input representations, and then updated during the training, that is, the output of the current layer is fed to the next layer.
Secondly, description texts are represented by the knowledge module, which is a light model outside the backbone model, and aims to represent the long description text separately.
Moreover, each knowledge layer is selectively connected to one backbone layer, to enhance corresponding text representation. Note that, the description representations are updated by the knowledge module layer by layer, but kept fixed when feed to the backbone layer.
Finally, two entity/description-related auxiliary tasks, namely entity/description enhancement and entity enhancement/pollution task, are designed to narrow the semantic gaps among the representations of texts, entities and descriptions.
% 5. 实验表明。。
We conduct experiments on two entity-related tasks, i.e., entity typing and relation classification, and two common NLP tasks, i.e., sentiment analysis and extended exact match. The experimental results show that \ourmodel~significantly outperforms the baseline models and gets SOTA on several datasets.

\section{Related Work}\label{related_work}
% pre-training stage, fine-tuning stage, +2021,2022
Some works have explored injecting entity or entity description into the pre-trained language models. Some of them include knowledge in the pre-training stage by designing pre-training tasks, while others introduce knowledge directly in the fine-tuning stage.

\textbf{In the pre-training stage.}
% 加个总领句，这种类别方法的通用做法：设计相关预训练任务
% ERNIE-THU
ERNIE-THU~\citep{zhang2019ernie} uses static entity embeddings separately learned from KGs. It first obtains all entity embeddings by TransE~\citep{AntoineBordes2013TranslatingEF}, links the named entity mentions in the text to entities in KGs, and adds the linked entity embeddings to the corresponding mention positions. Besides, it designs pre-training objectives by randomly masking some of the named entity alignments and asking the model to select appropriate entities from KGs to complete the alignments.
% KnowBert
Same to ERNIE-THU, KnowBert~\citep{peters2019knowledge} incorporates an integrated entity linker in their model and adopts end-to-end training.
% KEPLER
KEPLER~\citep{wang2021kepler} encodes entity descriptions by PLMs as the representations of entities and trains these entity representations by conventional knowledge embedding methods. The model is pre-trained by MLM and this KE objective.
% LUKE
In addition to the masked language model (MLM)~\citep{devlin2019bert}, LUKE~\citep{yamada2020luke} randomly masks tokens and entities and then recover them by training the RoBERTa to predict the tokens and the original form of the masked entities in KGs. 
It provides entity identifier ``[MASK]'' as additional input, and designs entity-aware self-attention to better use the entity identifier embedding.
% WKLM
WKLM~\citep{xiong2019pretrained} designs a pre-training task, which randomly replaces some of the entity names in the input text and asks the model to predict whether an entity name is replaced. 
K-Adapter~\citep{wang2021k} designs two adapters as a plug-in, which is pre-trained by relation classification and dependency relation prediction task.
% takes RoBERTa as the backbone and 

\textbf{In the fine-tuning stage}
KT-attn~\citep{xu2021does} appends entities and entity descriptions directly to the original input text in the fine-tuning stage and designs an attention matrix to avoid computation resource costs induced by descriptions. It also compares with knowledge as text and knowledge as embedding methods.

% The difference between the works mentioned above is ...
Our work is different from the works mentioned above. 
Firstly, both entities and entity descriptions are explicitly introduced to the fine-tuning stage.
Secondly, descriptions are processed by the knowledge module, a lighter model, to avoid the impacts induced by these long texts. Besides, the backbone and knowledge module is connected layer-to-layer.
Although it appears similar but is different from ~\citep{wang2021k}, where hidden states flow from the backbone to the pre-trained adapters. It is naturally a method of knowledge introduction by updating the weights of the models, whereas, the hidden states of \ourmodel~flow from the knowledge module to the backbone model for enhancement.

\section{Our Method}
\begin{figure}[t]
\centering
\includegraphics[width=.75\columnwidth]{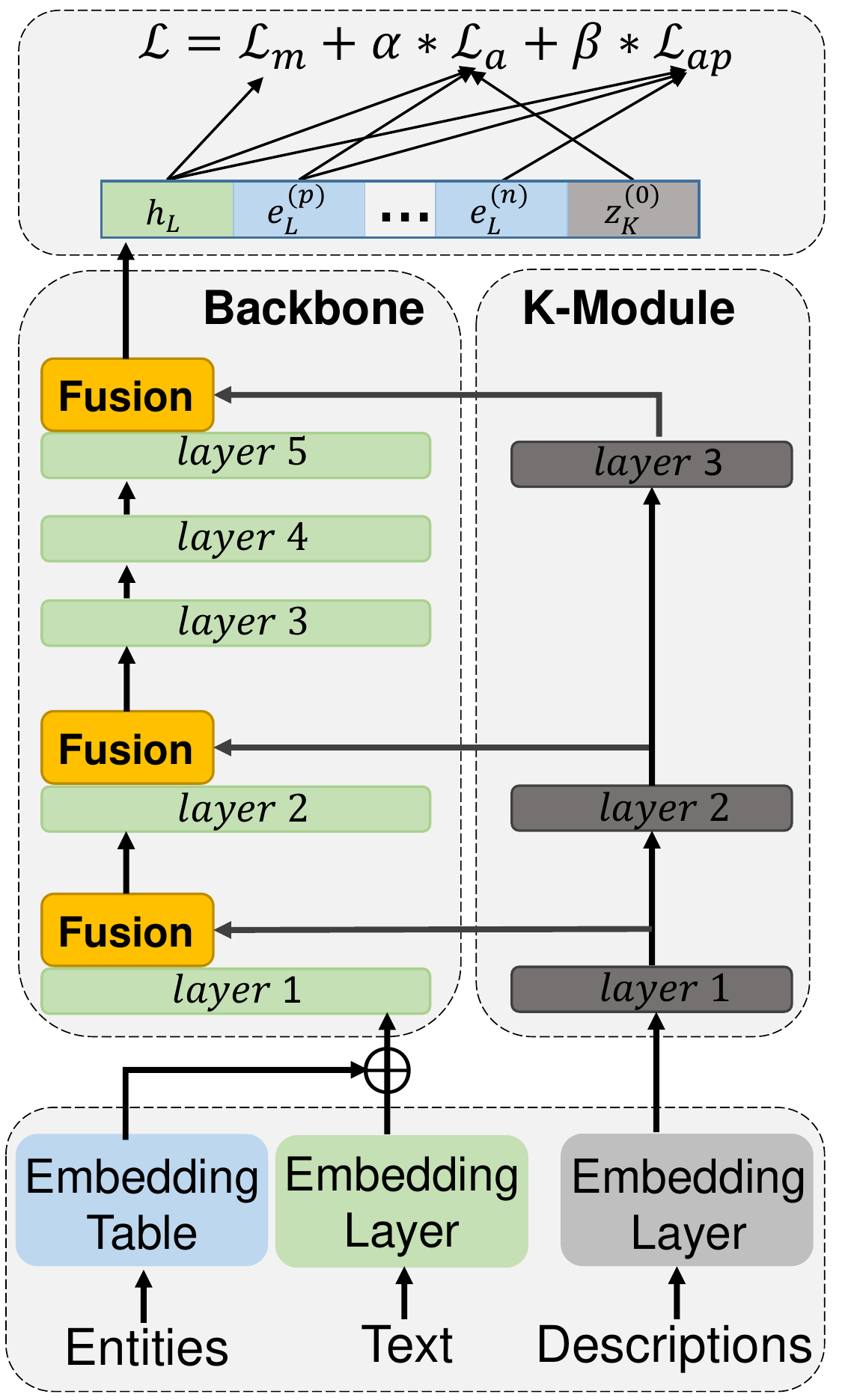}
\caption{Overview of \ourmodel~($L=5, K=3$, the connected layer is $1, 2, 5$). }
\label{framework}
\end{figure}

In this section, we present the overall framework of \ourmodel, as shown in Figure~\ref{framework}. It is composed of an input layer converting input items to vectors (Section~\ref{input_layer}), a backbone model processing text (Section~\ref{backbone_model}), a knowledge module processing descriptions (Section~\ref{knowledge_module}), a fusion module builds layer-wise connections between the layers of the backbone and knowledge module (Section~\ref{fusion_module}), and a prediction layer computing the probability distribution of target classes (Section~\ref{prediction_layer}).

%%%%%%%%%%%%%%%%%%%%%%%%%%%
%%%%%%%%%%%%%%%%%%%%%%%%%%%

\subsection{Input Layer}\label{input_layer}
The input of \ourmodel~includes the original text, entities, and descriptions. The text is fed into the embedding layer of the backbone model, where token embedding, position embedding, and segment embedding are added together.
The embeddings of entities are lookup from the entity embeddings table, which is pre-trained by entity-related pre-training tasks.
The description is tokenized and then fed into the embedding layer of the knowledge module (K-module). 
To be specific, given the input sentence $S$, we recognize all the entities by entity linker, it will output the mention span in the input and the entities in the Wikidata, and then we associate each entity with its description. 
After that, we tokenize $S$ into subword sequence $X=\{x_1, \dots, x_m\}$, where $m$ is the maximum sequence length of the text\footnote{We pad zeros to keep the dimension.}. Then, we get its embeddings $\mathbf{X}= \{\mathbf{x}_1, \dots, \mathbf{x}_m\}$ by the backbone embedding layer. For each description $D$,  we tokenize it into subword sequence $U=\{u_1, \dots u_n\}$, where $n$ is the maximum sequence length of the descriptions. Then, we get its embedding $\mathbf{U}=\{\mathbf{u}_1, \dots \mathbf{u}_n\}$ by the knowledge module embedding layer. Besides, entity embeddings $\mathbf{e}$ are obtained from the entity embedding table $\mathbf{v} \in \mathbb{R}^{|V| \times d_1}$, where $|V|$ is the entity vocabulary size, $d_1$ is the dimension size of entity embeddings.

\subsection{Backbone Model}\label{backbone_model}
The backbone model is responsible for capturing semantic representation from the original input tokens. 
It is a prevalent PLMs, e.g., BERT and RoBERTa, stacking $L$ backbone layers, and we exclude a comprehensive description of this module and refer readers to ~\citep{devlin2019bert} and ~\citep{liu2019roberta} for details.
In our setting, Transformer~\citep{vaswani2017attention} encoder is used, it takes the embeddings $\mathbf{X}= \{\mathbf{x}_1, \dots, \mathbf{x}_m\}$ as input and computes layer-wise representation. The output of current layer is fed into the next layer,
\begin{equation}
    \mathbf{h}_i = \text{Transformer}_i(\mathbf{h}_{i-1}),  % 这儿应该是layer级别的
\label{backbone}
\end{equation}
where $\mathbf{h}_i \in \mathbb{R}^{m \times d_2}, i \in \mathbb{N}^{+}, i \in [1, L]$ is the representation of the text in the $i$-th backbone layer, and $\mathbf{h}_0 = \mathbf{X}$. $\text{Transformer}_i$ refers to the $i$-th layer, $d_2$ is the dimension size of the backbone model.

\subsection{Knowledge Module}\label{knowledge_module}
% 负责XXX包括XXX输入XXX计算XXX
The knowledge module is responsible for capturing the knowledge-related representations of entity descriptions.
It is a light PLMs, that stacks $K$ knowledge layers, outside the backbone model as an external plugin to process the long text. 
In our setting, the pre-trained DistilBERT~\citep{sanh2019distilbert} is used, and its parameters are frozen. 
The knowledge module takes the embeddings of entity descriptions as input, and it updates the hidden states of the descriptions layer by layer, 
\begin{equation}
    \mathbf{z}_k = \text{Knowledge}_k(\mathbf{z}_{k-1}),
\label{knowledge}
\end{equation}
where $\mathbf{z}_k \in \mathbb{R}^{n \times d_3}, k \in \mathbb{N}^{+}, k \in [1, K]$ is the representation of the description text in the $k$-th knowledge layer, and $\mathbf{z}_0 = \mathbf{U}$. $\text{Knowledge}_k$ refers to the $k$-th layer of the knowledge module, $d_3$ is the dimension size of the knowledge module.

\subsection{Fusion Module}\label{fusion_module}
% 负责，设计，输入
Since different models produce the text, entities and descriptions embeddings with different semantic spaces. The fusion module is responsible for narrowing the semantic gaps, fusing the knowledge-related information into the input representation, and outputting an entity/description enhanced text representation.
Motivated by~\citep{yang2021graphformers}, instead of fusing the final hidden state, we build a layer-wise connection between the backbone and knowledge layers, to achieve deeper integration.

It takes the text representation $\mathbf{h}$, entity embedding $\mathbf{e}$ and description representation $\mathbf{z}$ as input. Since the number layer $K$ of the knowledge module is always less than the number layer $L$ of the backbone layer, some of the backbone layers are connected while others are not. Therefore, an alignment is needed to determine which backbone layer is connected.
For connected layers, both entity embedding and corresponding layer-wise representation of descriptions are concatenated to the text representation, and then fed to the next backbone layer for enhanced text representation. 
Note that, $\mathbf{z}_k$ is only used to enhance $\mathbf{h}$. Formally,
\begin{equation}
\label{with_des}
\begin{aligned}
 &\mathbf{h}'_{i-1} = \mathbf{h}_{i-1} \mid\mid \mathbf{e}_{i-1} \mid\mid f(\mathbf{z}^{(0)}_{k-1}), \\
&\mathbf{h}_i, \mathbf{e}_i, \_ = \text{Transformer}_i(\mathbf{h}'_{i-1}),  \\
&\mathbf{z}_k = \text{Knowledge}_k (\mathbf{z}_{k-1}),
\end{aligned}
\end{equation}
where $f$ is a linear function to align dimension, $\mathbf{z}^{(0)}_{k}$ is the vector in the first position of description, i.e., ``[CLS]'', output by the $k$-th knowledge layer. 
For layers without connection, the entity embedding is concatenated to the text representation, and then fed to the next backbone layer for enhancement, 
\begin{equation}
\label{without_des}
\begin{aligned}
 &\mathbf{h}'_{i-1} = \mathbf{h}_{i-1} \mid\mid \mathbf{e}_{i-1},    \\
&\mathbf{h}_i, \mathbf{e}_i = \text{Transformer}_i(\mathbf{h}'_{i-1}).
\end{aligned}
\end{equation}
For example, as depicted in Figure~\ref{framework}, the backbone model has five layers and the knowledge module has three layers. The shown alignment is that the knowledge layer is connected to the first, second, and fifth backbone layer.

\subsection{Prediction Layer}\label{prediction_layer}
The prediction layer comprises linear layers to map the representation over probability distributions.

\textbf{Main task.} 
The vector of entity identifier $\mathbf{h}_L^{(I)}$ (detailed in Section~\ref{experiments})  is used as the final representation to compute the probability distribution, $\hat{p} = W_1 \cdot \mathbf{h}_L^{(I)} + b_1 $ . With the given probabilities, cross-entropy loss function is adopted to compute the loss of the main task, %$\mathcal{L}_m = - \frac{1}{Y} \sum_{y \in Y}  y \cdot \log(\hat{p})$.
\begin{equation}
\begin{aligned}
% & \hat{p} = W_1 \cdot h_L^{I} + b_1  \\
& \mathcal{L}_m = - \frac{1}{Y} \sum_{y \in Y}  y \cdot \log(\hat{p}).
\end{aligned}
\end{equation}

% auxiliary tasks
\textbf{Auxiliary tasks.} Since the vectors of entities, texts and descriptions are obtained from different models, there have different semantic gaps. To shrink these gaps, motivated by ~\citep{zhao2022kesa}, where sentiment words are used to construct enhanced and polluted sentence representation, we design two auxiliary tasks.  
The first auxiliary task is  \textbf{entity/description enhancement task}, which is pretty similar to the main task, except that the vector $\mathbf{h}_L^{(E)}$ of target entity or sentence is enhanced with the knowledge representations, $\mathbf{h}_{(a)} = \mathbf{h}_L^{(E)} + \mathbf{e}^{(p)}_L + \mathbf{z}_{K}^{(0)}$. Then, is is used as the final representation to compute the probability distribution over the target classes, $\hat{p} = W_2 \cdot h_{(a)} + b_2$. Therefore, the loss of the first auxiliary task is, 
\begin{equation}
\begin{aligned}
    % & h_{(a)} = h_L^{(E)} + e^{(p)}_L + z_{K}^{(0)}  \\
    % & \hat{p} = W_2 \cdot h_{(a)} + b_2 \\
    \mathcal{L}_a = - \frac{1}{Y}  \sum_{y \in Y}  y \cdot \log(\hat{p}). 
\end{aligned}
\end{equation}
The second auxiliary task is \textbf{entity enhancement/pollution task}, where the text representation is enhanced by the representation of its associated entity, i.e., $\mathbf{g}_{(a)} = \mathbf{e}^{(p)}_L + \mathbf{h}_L^{(E)}$ or polluted by randomly sampled ones, i.e., $\mathbf{g}_{(p)} = \mathbf{e}^{(n)}_L + \mathbf{h}_L^{(E)}$, and the model is asked to distinguish them $\hat{c} = W_3 \cdot (\mathbf{g}_{(a)} \mid\mid \mathbf{g}_{(p)} ) + b_3$. Therefore, the loss of the second auxiliary task is, %$\mathcal{L}_{ap} = - \frac{1}{C} \sum_{c \in C} c \cdot \log(\hat{c})$.
\begin{equation}
\begin{aligned}
    % & g_{(a)} = e^{(p)}_L + h_L^{(E)}\\
    % & g_{(p)} = e^{(n)}_L + h_L^{(E)} \\
    % & \hat{c} = W_3 \cdot (g_{(a)} \mid\mid g_{(p)} ) + b_3  \\
    \mathcal{L}_{ap} = - \frac{1}{C} \sum_{c \in C} c \cdot \log(\hat{c}),
\end{aligned}
\end{equation}
where $\mid\mid$ refers to concatenation operation. $Y$ is the label set of the main task, and $C$ is the label set indicating the position index of the positive entity. $W_1, b_1, W_2, b_2, W_3, b_3$ are model parameters, $\mathbf{e}^{(p)}_L, \mathbf{e}^{(n)}_L$ refer to the representation of the positive and negative entities, respectively. $\mathbf{z}_{K}^{(0)}$ is the vector in the first position of the last knowledge layer. $\mathbf{h}_L^{(I)}$ and $\mathbf{h}_L^{(E)}$ is the vector in the position $(I), (E)$ of the last backbone layer, and $(I), (E)$ index the position of the entity identifier and entity special token, respectively.
It will be detailed in Section~\ref{experiments}.

The total loss is a weighted sum of the above three losses, $\mathcal{L} = \mathcal{L}_m + \alpha * \mathcal{L}_a + \beta * \mathcal{L}_{ap}$, 
% \begin{equation}
%     \mathcal{L} = \mathcal{L}_m + \alpha * \mathcal{L}_a + \beta * \mathcal{L}_{ap}
% \label{total_loss}
% \end{equation}
where $\alpha > 0$ and $\beta > 0$ are loss coefficients.

\section{Experiments}
\label{experiments}
This section presents the implementation details and the results of several NLP tasks.
The statistics of these datasets are shown in Table~\ref{openentity_statistics}.
We use LUKE~\cite{yamada2020luke} as the backbone model and DistilBERT as the knowledge module. LUKE is based on the large version of RoBERTa ($L=24, d_2=1024$) and DistilBERT is a distilled BERT with $K=6, d_1 = 768$.
We extract descriptions from Wikidata\footnote{\url{https://www.wikidata.org/w/api.php}}, and entity embedding table from the pre-trained LUKE checkpoints\footnote{\url{https://huggingface.co/studio-ousia/luke-large}}, which contains 500,000 entities.
Positive entities are recognized by the entity linker, while negative entities are randomly sampled from entity vocabulary. 
Both baselines and our method share the same training parameter for fairness.
Note that, we run the experiments several times and report the average results except for \figer~dataset. 
Please refer to the Appendix~\ref{implementation_details} for more implementation details.
The source code is available at XXX (we will release all the code when the paper is accepted).

% %%%%%%%%%%%%%%%%%%%%%%%%%%%%%%%%%%%%%%%%%%%%%%%%%%%%
\begin{table}[h]
\centering
\resizebox{\columnwidth}{!}{
\begin{tabular}{llllc}
\hline 
Dataset& Train&  Dev&    Test&   \#Types \\
\hline
\openentity&    1,998&  1,998&    1,998&   9   \\
\figer&   2,000,000&  10,000&  563&    113  \\
\fewrel&    8,000&   16,000&  16,000&    80  \\
\tacred&    68,124& 22631&  15,509& 42  \\
\sst&   67,349&   872&  --&   2\\
\eem&   405,482&   101,370&   --&   2\\
\hline
\end{tabular}
}
\caption{The statistics of \openentity, \figer, \fewrel, \tacred, \sst~and \eem~datasets.}
\label{openentity_statistics} 
\end{table}
% %%%%%%%%%%%%%%%%%%%%%%%%%%%%%%%%%%%%%%%%%%%%%%%%%%%%

\subsection{Knowledge-orientated Tasks}
We first conduct experiments on knowledge-oriented tasks, i.e., entity typing and relation classification. Baselines are described in section~\ref{related_work}.

\subsubsection{Entity Typing}
Entity typing is the task of predicting the types of an entity given its sentence context. Here we use Open Entity~\cite{choi2018ultra} and FIGER~\cite{ling2015design} datasets, following the split setting as~\cite{zhang2019ernie, wang2021k}. 
To fine-tune our models for entity typing, following the setting of ~\cite{yamada2020luke}, we modify the input token sequence by adding the special token ``[ENTITY]'' before and after a certain entity, and providing entity identifier ``[MASK]'' along with the input. 
The representation of entity identifier ``[MASK]'' is adopted to perform classification, and the first ``[ENTITY]'' special token representation is used as text representation. 
It is treated as a multiple labels classification problem, and binary cross-entropy loss is used to optimize the model. 
Following the same evaluation criteria used in the previous works, for \openentity, we evaluate the models using micro precision, recall and F1, and adopt the micro F1 score as the final metric. For \figer, 
we adopt accuracy, macro F1, and micro F1 scores for evaluation.

%%%%%%%%%%%%%%%%%%%%%%%%%%%%%%%%%%%%%%%%%%%%%%%%%%%%
\begin{table}%[t]
\centering
\resizebox{\columnwidth}{!}{
\begin{tabular}{l|l|l|c}
\hline 
Model&   Prec.& Rec.&  Mi-F1 \\ 
\hline
BERT$_\text{base}$&   76.4&   71.0&   73.6\\
ERNIE-THU~\cite{zhang2019ernie}&  78.4&   72.9&   75.6\\
KnowBERT~\citep{peters2019knowledge}&   78.6&   73.7&  76.1\\
\hline
RoBERTa$_\text{large}$&    77.6&   75.0& 76.2\\
K-Adapter~\citep{wang2021k}&  79.0&  76.3&   77.6\\
LUKE~\citep{yamada2020luke}&   79.9&   76.6&  78.2\\
\hline
RoBERTa$_\text{large}^*$&   78.3&   74.4&   76.3\\
K-Adapter$^*$&  78.0&  76.3&  77.0   \\
LUKE$^*$&   80.8&  74.7&  77.6    \\
LUKE+Adapter&   78.3&  76.1&  77.4\\
\textbf{\ourmodel}&    80.3&   75.9&   \textbf{78.1}\\
% \ourmodel$_{-a}$&   80.43&  74.61&  77.41 \\
% \ourmodel$_{-b}$&    80.09& 75.3&  77.62\\
% \ourmodel$_{-a-b}$&   79.92&    75.30&  77.54 \\

\hline
\end{tabular}
}
\caption{Results of entity typing on the \openentity~dataset. $^*$ refers to reproduced results.}
\label{openentity} 
\end{table}
%%%%%%%%%%%%%%%%%%%%%%%%%%%%%%%%%%%%%%%%%%%%%%%%%%%%

%%%%%%%%%%%%%%%%%%%%%%%%%%%%%%%%%%%%%%%%%%%%%%%%%%%%
\begin{table}[t]
\centering
\resizebox{\columnwidth}{!}{
\begin{tabular}{l|l|l|c}
\hline 
Model&   Acc& Ma-F1&  Mi-F1 \\ 
\hline
BERT$_\text{base}$&   52.0&  75.2&   71.6\\
ERNIE-THU~\cite{zhang2019ernie}&   57.2&   75.6&   73.4\\
WKLM~\cite{xiong2019pretrained}&   60.2&   82.0&   77.00\\
\hline
RoBERTa$_\text{large}$&    56.3& 82.4&   77.8\\
K-Adapter~\citep{wang2021k}&  61.8&    84.9&   80.5\\
\hline
RoBERTa$_\text{large}^*$&   54.9&   81.6&   77.1\\
LUKE$^*$&   57.4&   \textbf{82.1}&   78.1\\

\textbf{\ourmodel}&    \textbf{60.6}&    77.7&   \textbf{78.8}\\
% \ourmodel$_{-a}$&  59.9&    81.1&   78.4\\
% \ourmodel$_{-b}$&  60.4&    76.7&   78.6\\
% \ourmodel$_{-a-b}$&  59.0& 76.3&   78.1\\

\hline
\end{tabular}
}
\caption{Results of entity typing on the \figer~dataset  (maximum sequence length is reduced from 256 to 128).}
\label{figer} 
\end{table}
%%%%%%%%%%%%%%%%%%%%%%%%%%%%%%%%%%%%%%%%%%%%%%%%%%%%

\textbf{Results}
The results on \openentity~and \figer~dataset are presented in Table~\ref{openentity} and ~\ref{figer}, respectively. We can see that \ourmodel~outperforms the previous SOTA by 0.5 F1 points on \openentity. On \figer~dataset, it outperforms the reproduced RoBERTa and LUKE by 1.7 and 0.7 micor F1 points, respectively.
Besides, to demonstrate the effectiveness of our proposed model, we also reimplement LUKE+Adapter, where the two adapters pre-trained by K-Adapter~\cite{wang2021k} are transferred to the LUKE model. We find that, with the plugin of the two adapters, there are no expected gains, but drops of 0.2 F1 points on the \openentity~dataset. We attribute these results to the semantic spaces of the two, namely LUKE and adapters, being different. We think it is necessary to narrow this semantic gap, and the ablation study in Section~\ref{ablation_study} confirms our view.

\subsubsection{Relation Classification}
Relation classification is the task of determining the relation between the given head and tail entities in a sentence. Here we use TACRED~\cite{zhang2017position} and FewRel~\cite{han2018fewrel} datasets, following the split setting as~\cite{zhang2019ernie, wang2021k}.
Following~\cite{yamada2020luke}, we modify the input token sequence by adding the special token ``[HEAD]'' before and after the first entity, adding ``[TAIL]'' before and after the second entity, and adding two entity identifiers ``[MASK]'' as additional input. 
The representations of entity identifiers are concatenated to perform relation classification, and the token representations of the first special token ``[HEAD]'' and ``[TAIL]'' are concatenated to represent the original text.
We evaluate the models using micro precision, recall and F1, and adopt micro F1 score as the final metric to represent the model performance as in previous works.

% %%%%%%%%%%%%%%%%%%%%%%%%%%%%%%%%%%%%%%%%%%%%%%%%%%%%
\begin{table}%[t]
\centering
\resizebox{\columnwidth}{!}{
\begin{tabular}{l|l|l|c}
\hline 
Model&   Prec.& Rec.&  Mi-F1 \\ 
\hline
BERT$_\text{base}$&   85.1&  85.1&   84.9\\
ERNIE-THU~\cite{zhang2019ernie}&  88.5&   88.4&   88.3\\
\hline
RoBERTa$_\text{large}^*$&   88.8&   88.8&   88.8\\
LUKE$^*$&   89.4&  89.4&  89.4\\

\textbf{\ourmodel}&  90.3&  90.3&  \textbf{90.3}\\
% \ourmodel$_{-a}$&  90.16&   90.16&  90.16\\
% \ourmodel$_{-b}$&   89.89&  89.89&  89.89  \\
% \ourmodel$_{-a-b}$&  89.94& 89.94&  89.94\\

\hline
\end{tabular}
}
\caption{Results of entity typing on the \fewrel~dataset.}
\label{fewrel} 
\end{table}
% %%%%%%%%%%%%%%%%%%%%%%%%%%%%%%%%%%%%%%%%%%%%%%%%%%%%

% %%%%%%%%%%%%%%%%%%%%%%%%%%%%%%%%%%%%%%%%%%%%%%%%%%%%
\begin{table}%[t]
\centering
\resizebox{\columnwidth}{!}{
\begin{tabular}{l|l|l|c}
\hline 
Model&   Prec.& Rec.&  Mi-F1 \\
\hline
BERT$_\text{base}$&   67.2&  64.8&   66.0\\
ERNIE-THU~\cite{zhang2019ernie}&  70.0&   66.1&   67.97\\
KnowBERT~\citep{peters2019knowledge}&   71.6&   71.4&   71.5\\
\hline
RoBERTa$_\text{base}$&    70.4&   71.1&   70.7\\
KEPLER-Wiki~\cite{wang2021kepler}& 71.5&   72.5&   72.0\\
\hline
RoBERTa$_\text{large}$&  70.2&   72.4&  71.3\\
K-Adapter~\citep{wang2021k}&  70.1&    74.0&   72.0\\
LUKE~\citep{yamada2020luke}&   70.4&   75.1&   72.7\\
\hline
LUKE$^*$&   71.2&  72.2&  71.7   \\

\textbf{\ourmodel}&  71.3&  73.7&  \textbf{72.5}\\
% \ourmodel$_{-a}$&  70.80&   72.54&  71.65\\
% \ourmodel$_{-b}$&  72.25&   72.27&  72.14\\
% \ourmodel$_{-a-b}$&  72.73& 71.05&  71.83\\

\hline
\end{tabular}
}
\caption{Results of relation classification on \tacred.}
\label{tacred} 
\end{table}
% %%%%%%%%%%%%%%%%%%%%%%%%%%%%%%%%%%%%%%%%%%%%%%%%%%%%

\textbf{Results}
% 1. 结果如表所示
The results on \fewrel~and \tacred~are shown in Table ~\ref{fewrel} and ~\ref{tacred}, respectively. Notably, the gap between the original reported results in LUKE and the reproduced results may probably be because of different maximum sequence lengths, i.e., from 512 to 256, open-source library, i.e., from AllenNLP to HuggingFace, and reports, i.e., from the best to average results. 
Compared with the previous best-published models, \ourmodel~achieves an improvement of 0.9 and 0.8 F1 points, respectively, demonstrating the usefulness of the representations of entities and entity descriptions and the effectiveness of our designed framework.

\subsection{Common Tasks}
\cite{zhang2019ernie, wang2021kepler, xu2021does} show that common tasks may not require external knowledge, which may harm the language model's representation to some extent.
To test \ourmodel, we conduct experiments on two common tasks, including sentence-level sentiment analysis and extended exact match tasks.

%%%%%%%%%%%%%%%%%%%%%%%%%%%%%%%%%%%%%%%%%%%%%%%%%%%%
\begin{table}[t]
\centering
\resizebox{.92\columnwidth}{!}{
\begin{tabular}{l|l}
\hline 
Model&   ACC \\ 
\hline

BERT$_\text{base}$&  93.00\\
ERNIE-THU~\cite{zhang2019ernie}&  93.50 \\
KT-attn$_\text{bert-base}^*$&    93.33\\
\hline
RoBERTa$_\text{base}$&    94.72\\
KEPLER-Wiki~\cite{wang2021kepler}& 94.50\\
KT-attn$_\text{roberta-base}$~\citep{xu2021does}&      94.84 \\
KT-attn$_\text{roberta-base}^*$&       94.72\\

\hline
RoBERTa$_\text{large}$&  96.22\\
RoBERTa$_\text{large}^*$&   96.10 \\
KT-attn$_\text{roberta-large}$~\citep{xu2021does}&    96.44\\
KT-attn$_\text{roberta-large}^*$&   96.44 \\
\textbf{\ourmodel}&  \textbf{96.90}\\
% \ourmodel$_{-a}$&  96.33\\
% \ourmodel$_{-b}$&  96.10\\
% \ourmodel$_{-a-b}$&  96.44\\

\hline
\end{tabular}
}
\caption{Results of sentiment analysis on the \sst~dataset. $^*$ refers to reproduced results. }
\label{sst} 
\end{table}
% %%%%%%%%%%%%%%%%%%%%%%%%%%%%%%%%%%%%%%%%%%%%%%%%%%%%

% %%%%%%%%%%%%%%%%%%%%%%%%%%%%%%%%%%%%%%%%%%%%%%%%%%%%
\begin{table}[t]
\centering
\resizebox{.85\columnwidth}{!}{
\begin{tabular}{l|c|c}
\hline 
Model&   ROC AUC& PR AUC\\ 
\hline

BERT$_\text{base}^*$&  85.60& 90.64  \\
KT-attn$_\text{bert-base}^*$&    86.27&    91.02\\
\hline
RoBERTa$_\text{base}^*$&   86.08&   90.94\\
KT-attn$_\text{roberta-base}^*$&    86.87&  91.38\\
\hline
RoBERTa$_\text{large}^*$&  87.36&  91.82\\
KT-attn$_\text{roberta-large}^*$&  \textbf{88.29}&  92.46\\

\hline
\textbf{\ourmodel}&  87.90&  92.27\\
% 等有计算资源的时候，对EEM使用TAGME做EL
% \ourmodel$_{-a}$& 87.73&    92.11\\
% \ourmodel$_{-b}$&  87.67&   92.06\\
% \ourmodel$_{-a-b}$&  87.82& 92.17\\

\hline
\end{tabular}
}
\caption{Results on the \eem~dataset.}
\label{eem} 
\end{table}
%%%%%%%%%%%%%%%%%%%%%%%%%%%%%%%%%%%%%%%%%%%%%%%%%%%%

%%%%%%%%%%%%%%%%%%%%%%%%%%%%%%%%%%%%%%%%%%%%%%%%%%%%
\begin{table*}[t]
\centering
\resizebox{\textwidth}{!}{
\begin{tabular}{l|lll|ccc|lll|lll|l|cc}
\hline 
Dataset&    \multicolumn{3}{c|}{\openentity}&   \multicolumn{3}{c|}{\figer}& \multicolumn{3}{c|}{\fewrel}&   \multicolumn{3}{c|}{\tacred}&   \sst&      \multicolumn{2}{c}{\eem}  \\
Model&  Prec.& Rec.&  Mi-F1&      Acc& Ma-F1&  Mi-F1&     Prec.& Rec.&  Mi-F1&    Prec.& Rec.&  Mi-F1&   ACC&  ROC AUC& PR AUC \\
\hline
Baseline&   80.8&   74.7&   77.6&   57.4&  \textbf{ 82.1}&   78.1&   89.4&   89.4&   89.4&   71.2&   72.2&   71.7&   96.4&  \textbf{88.3}&  92.5\\
\hline
\ourmodel&    80.3&   75.9&   \textbf{78.0}&  \textbf{60.6}&   77.7&   \textbf{78.8}&   90.3&  90.3&  \textbf{90.3}& 71.3&  73.7&  \textbf{72.5}&  \textbf{96.9}&  87.9&  92.3\\
w/o a&   80.4&  74.6&  77.4&  59.9&   81.1&   78.4&  90.2&   90.2&  90.2&    70.8&   72.5&  71.7&  96.3&   87.7&    92.1\\
w/o b&    80.1& 75.3&  77.6&   -&   -&   -& 89.9&  89.9&  89.9&    72.3&   72.3&  72.1&  96.1& 87.7&   92.1\\
w/o a, b&   79.9&    75.3&  77.5&  -&   -&   -&  89.9&   89.9&  89.9& 72.7& 71.1&  71.8&  96.4&   87.8&   92.2\\

\hline
\end{tabular}
}
\caption{Ablation results of each auxiliary task.}
\label{ablation_results} 
\end{table*}
%%%%%%%%%%%%%%%%%%%%%%%%%%%%%%%%%%%%%%%%%%%%%%%%%%%%

\textbf{Sentence-level sentiment analysis} aims to predict the sentiment polarity of the given sentence.
We use \sst~dataset, obtained from General Language Understanding Evaluation (GLUE) benchmark~\citep{wang2018glue}, which collects several popular NLP tasks~\citep{PranavRajpurkar2016SQuAD1Q, RichardSocher2013RecursiveDM}. The entity identifier ``[MASK]'' is inserted and its representation is used for prediction.
While, the vector in the first position of the last backbone layer is adopted to compute the enhanced or polluted text representation, we evaluate it by accuracy (ACC).

\textbf{Extended exact match} is a kind of matching mode of search advertising in the search advertisements scene, which requires the user's search term, i.e., query, must exactly match the bid term, i.e., keyword. Therefore, the task is to determine whether the given query and keyword are exactly matched or not. It is a binary classification problem.
A private \eem~dataset from the Bing ads group is used, where an entity identifier is provided as an extra input and the vector in the first position is used to perform all classifications. We evaluate it by ROC AUC and PR AUC.

\textbf{Results}
Table~\ref{sst} shows the results on \sst. We can see that \ourmodel~outperforms all the baselines, and increases the accuracy of RoBERTa$_\text{large}$ and KT-attn by 0.8 and 0.45 accuracy points, respectively.
Table~\ref{eem} shows the results on \eem, it shows that \ourmodel~outperforms the RoBERTa$_\text{large}$ about 0.6 ROC AUC points and is comparable to KT-attn.
We claim that common tasks are not knowledge-intensive, but the reasonable use of knowledge can promote the representation of the language model, just as our human beings do.

\subsection{Ablation Study}
\label{ablation_study}
% 1. 本段要做啥，2. w/o表示啥，3.分析，4.结论
In this subsection, we analyze the impacts of external knowledge and auxiliary tasks, where \textbf{w/o a} refers to fine-tuning \ourmodel~without entity/description enhancement task, \textbf{w/o b} refers to removing entity enhancement/pollution task, \textbf{w/o a, b} refers to no auxiliary is adopted except entity and description representation.
As shown in Table~\ref{ablation_results}, \textbf{w/o a} is better than \textbf{w/o b} is some cases but worser in other cases. 
\ourmodel~is better than \textbf{w/o}, demonstrating the necessity of the auxiliary tasks and two auxiliary tasks can mutually enhance each other.
Moreover, when no auxiliary task is adopted, the ablation models suffer significant drops, about 0.3 to 0.8 points, demonstrating that the straightforward introduction of external representation may not be helpful or even harm the performance.
In summary, according to the results, we claim that when external representation is introduced, which may have a different semantic space from the backbone, auxiliary tasks that aim to \textbf{narrow semantic gap} are necessary. These results also explain that combining pre-trained adapters from K-Adapter with LUKE does not boost the performance.

\section{Analysis}

\subsection{Effects of Layer Alignment}
As described in subsection~\ref{fusion_module}, 
each knowledge layer is selectively connected with one backbone layer. Therefore, in this section, we analyze the impacts of different layer alignment, the results are shown in Table~\ref{layer_connection}. 
For the test four datasets, the differences between the best and worst results are 0.7, 0.1, 0.6, and 0.5, respectively, indicating that different layer alignment has significant impacts.
``last'' can achieve the top results on \openentity, \fewrel~and \tacred dataset, but obtain the worst results on \sst, demonstrating different layer fusion impacts different datasets. 
In most cases, ``first \& last'' can get a relatively solid result. 

% %%%%%%%%%%%%%%%%%%%%%%%%%%%%%%%%%%%%%%%%%%%%%%%%%%%%
\begin{table}[h]
\centering
\resizebox{\columnwidth}{!}{
\begin{tabular}{llllll}
\hline 
Layers&   \openentity&    \fewrel&    \tacred&    \sst&   avg.\\
\hline
last&  \underline{78.1}&   \underline{90.4} &  \underline{72.6}&   95.6&   84.18\\
first& 77.6&   90.3&   \underline{72.5}&   \underline{96.1}&   84.13\\
middle&   77.5&     90.3&  72.0&   \underline{96.1}&   83.98\\
first \& last& \underline{78.1}&   90.3&   72.4&   \underline{96.1}&   \textbf{84.23}\\
uniform&    \underline{78.2}&   \underline{90.4}&   72.0&   96.0&   84.15\\
\hline
\end{tabular}
}
\caption{Results under different layer alignments. ``last'', ``first'', ``middle'' refers to the last/first/middle $K$ backbone layers are connected, and ``first \& last'' refers to the first and last $K/2$ backbone layers are connected. ``uniform'' means that the backbone layers are connected to knowledge layer in a uniform interval. Underline indicates top ranked results.}
\label{layer_connection} 
\end{table}
% %%%%%%%%%%%%%%%%%%%%%%%%%%%%%%%%%%%%%%%%%%%%%%%%%%%%

% %%%%%%%%%%%%%%%%%%%%%%%%%%%%%%%%%%%%%%%%%%%%%%%%%%%%
\begin{table*}%[t]
\centering
\resizebox{.9\textwidth}{!}{
\begin{tabular}{lcccc||lcccc}
\hline 
$\alpha$&   \openentity&    \fewrel&    \tacred&    \sst& $\beta$&   \openentity&    \fewrel&    \tacred&    \sst\\
\hline
2.0& 76.8&  89.7&  71.7&   96.0   &2.0&   68.0&    90.0&    70.8&  95.7\\
1.0& \underline{77.5}& \underline{89.8}&   \underline{71.9}&   95.6   &1.0& 68.0&  90.1&   71.0&   95.5\\
0.5& \underline{77.5}& 89.6&   \underline{72.0}&   96.1   &0.5& 68.0&  \underline{90.4}&   72.0&   95.9\\
0.1& \underline{77.5}& \underline{89.7}&   71.7&   95.9   &0.1& 77.1&  \underline{90.2}&   71.6&   \underline{95.9}\\
0.05& 76.5&   \underline{89.9}&    \underline{72.0}&   \underline{96.2}   &0.05&  \underline{77.4}&    \underline{90.0}&   \underline{72.5}  & 95.8\\
0.01& 75.3&   89.6&    71.1&   \underline{96.1}   &0.01&  \underline{78.4}&     90.0&  \underline{72.3} &  \underline{96.0}\\
0.005& 73.6&  89.7&    71.5&   \underline{96.3}   &0.005&  \underline{77.7}&   89.7&  71.8  &  95.5\\
0.001& 72.2& 89.5& 70.1&   95.9   &0.001&  77.2&   89.9&  \underline{72.2}  &  \underline{96.4}\\
\hline
\end{tabular}
}
\caption{Results under different values of $\alpha$ and $\beta$. Underline indicates top ranked results.}
\label{loss_weights_a_b} 
\end{table*}
% %%%%%%%%%%%%%%%%%%%%%%%%%%%%%%%%%%%%%%%%%%%%%%%%%%%%

% %%%%%%%%%%%%%%%%%%%%%%%%%%%%%%%%%%%%%%%%%%%%%%%%%%%%
\begin{figure*}%[t]
\centering
\subfigure[\openentity]{\label{openentity_weights}
\includegraphics[width=0.49\columnwidth]{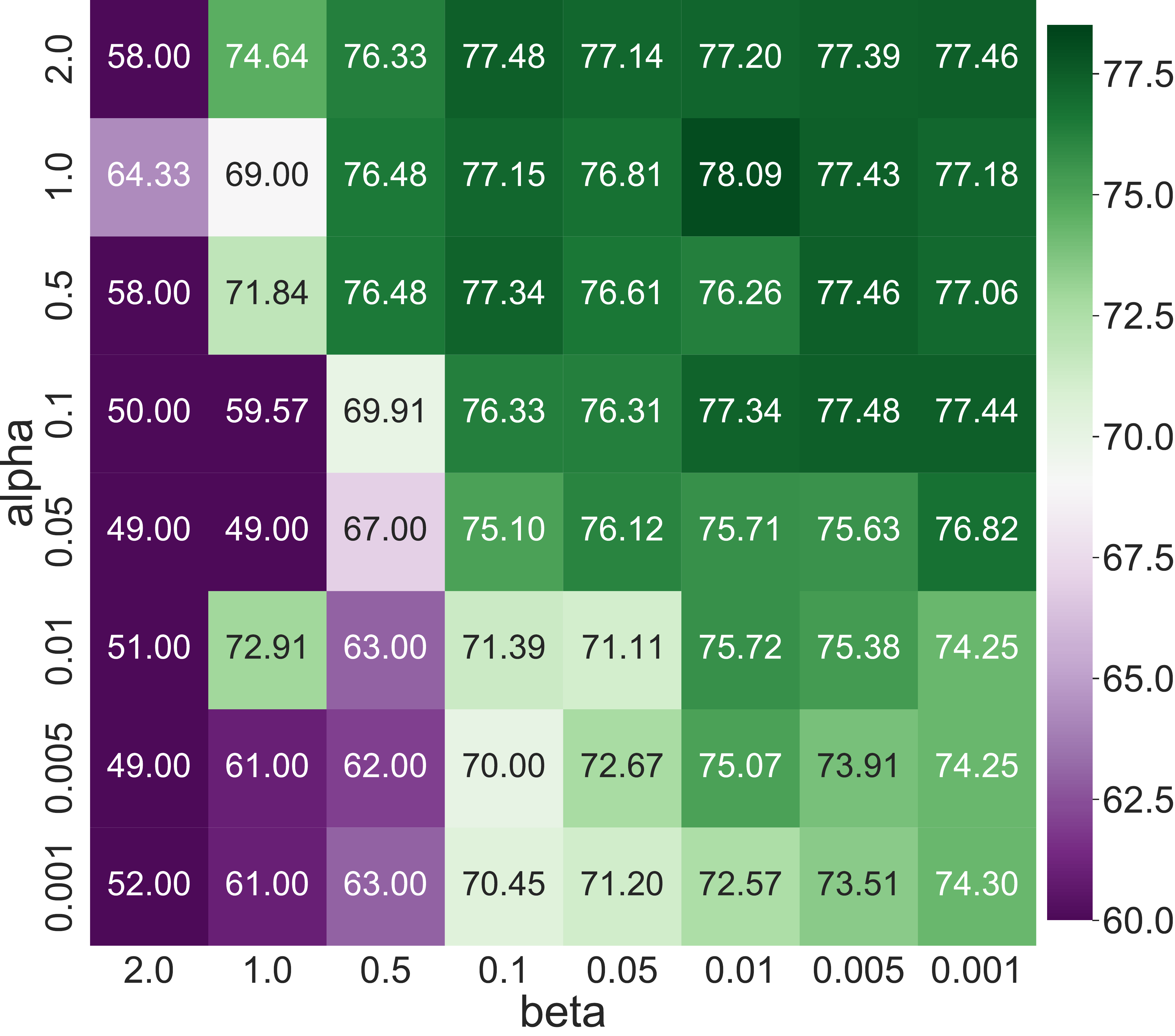}}
\subfigure[\fewrel]{\label{fewrel_weights}
\includegraphics[width=0.49\columnwidth]{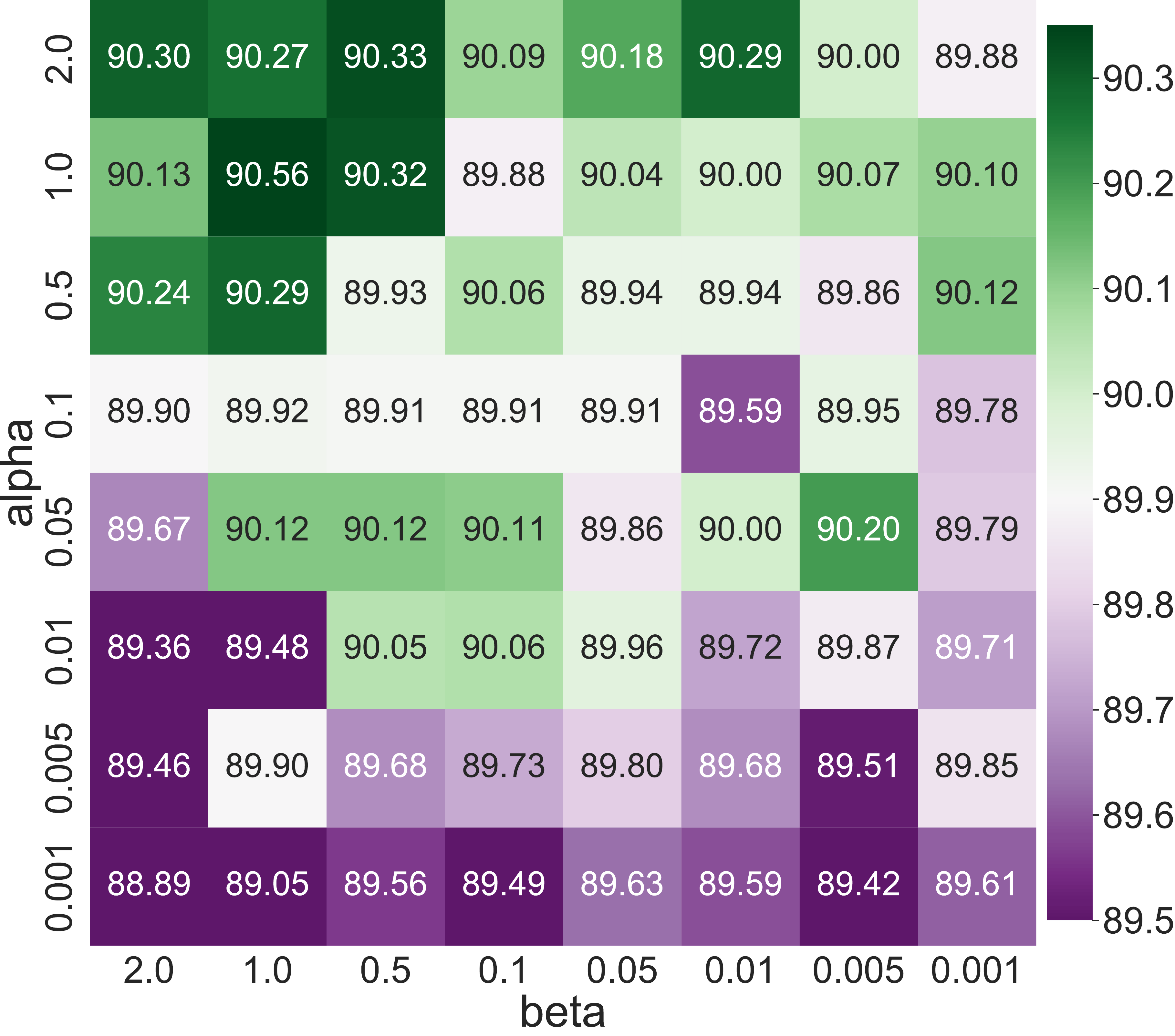}}
\subfigure[\tacred]{\label{tacred_weights}
\includegraphics[width=0.49\columnwidth]{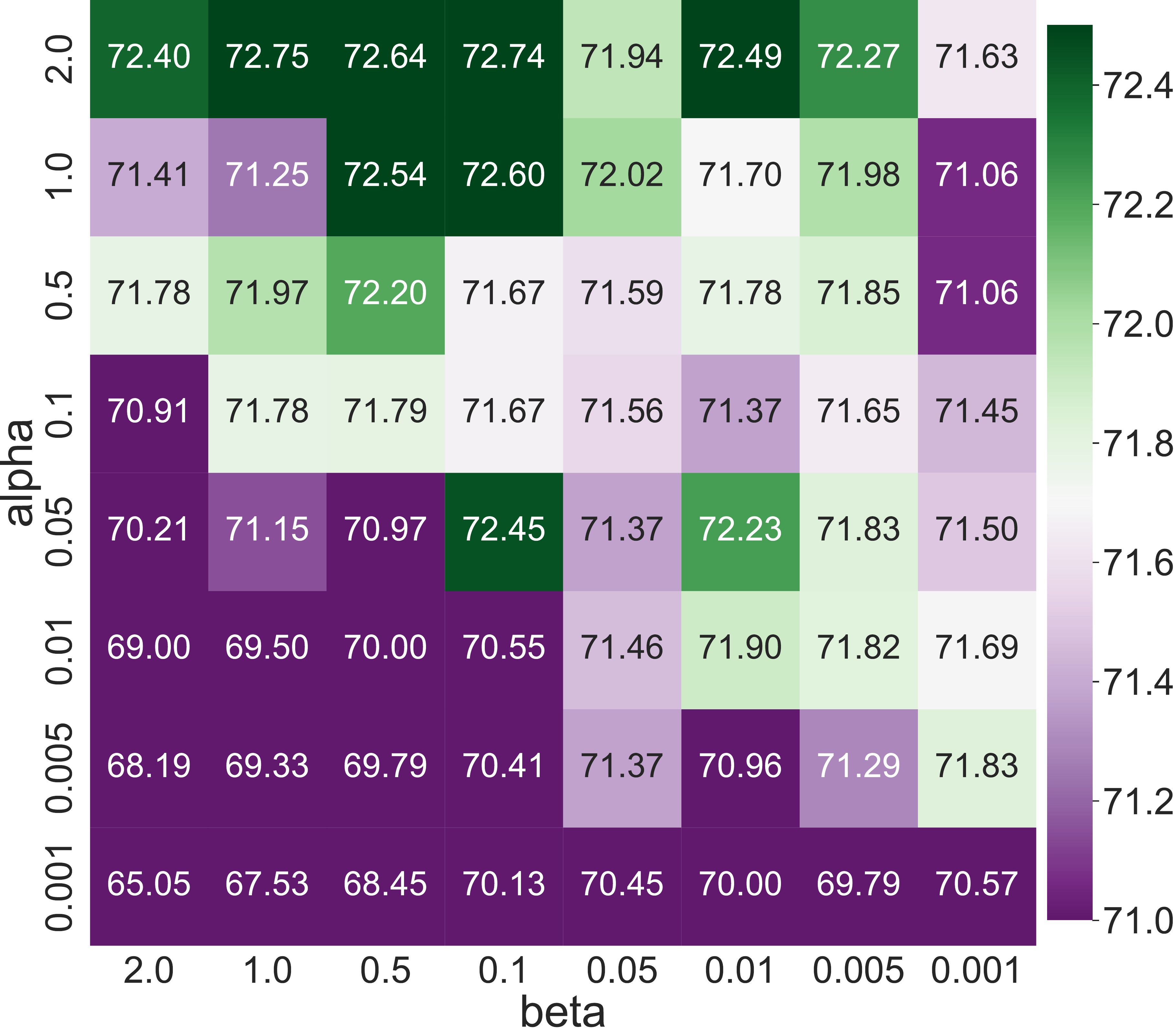}}
\subfigure[\sst]{\label{sst_weights}
\includegraphics[width=0.49\columnwidth]{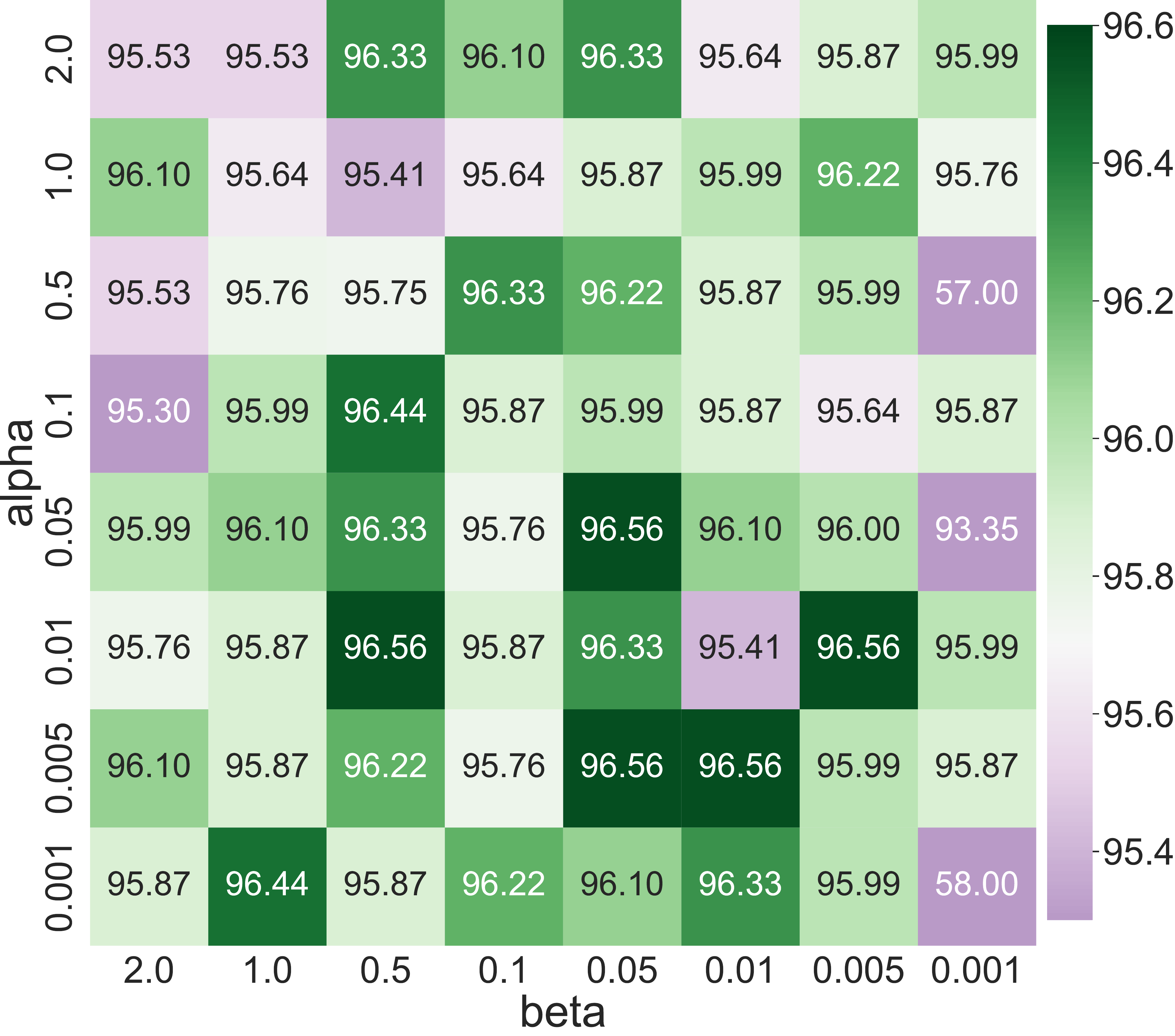}}
\caption{Impacts of loss coefficients.}
\label{loss_weights_both}
\end{figure*}
% %%%%%%%%%%%%%%%%%%%%%%%%%%%%%%%%%%%%%%%%%%%%%%%%%%%%

\subsection{Effects of Loss Coefficients}
In Section~\ref{prediction_layer}, we use $\alpha$ and $\beta$ coefficients to weight the two auxiliary task losses and then add it with the main loss. As reported in previous works~\cite{zhao2022kesa, chuang2022diffcse}, the auxiliary loss should have smaller weights to avoid domain the model's attention. 
Therefore, in this section, we search $\alpha$ and $\beta$ from \{2.0, 1.0, 0.5, 0.1, 0.05, 0.01, 0.005, 0.001\} to demonstrate this parameter's impacts, the results are shown in Table~\ref{loss_weights_a_b}. 
We can find that when we solely adopt the first task, i.e., the entity/description enhancement task, 1.0 to 0.05 is better for the three knowledge-related datasets, and 0.05 to 0.005 is better for \sst.
When we solely adopt the second task, namely the entity enhancement/pollution task, 0.01 to 0.005 is a better choice.
In summary, a relatively larger $\alpha$ and smaller $\beta$ are recommended in most cases.
Figer~\ref{loss_weights_both} shows the mutual impacts of the two auxiliary tasks. It shows that top results concentrate on the top of Figer~\ref{openentity_weights}, \ref{fewrel_weights} and \ref{tacred_weights}, whereas on the contrary for \sst. For \fewrel~and \tacred, the results in the top left corner are better, whereas for \openentity, that in the top right corner are better.

% \subsection{Effects of the Coverage}
% % 在general任务上是否有效
% % 只测试common task，把entity覆盖逐渐增加
% % 分两种情况，第一种test不变，train的coverage逐渐增加看看performance是不是越来越好；第二种情况，train保持固定，test的knowledge coverage是不是越高效果越好
% % 这个实验不在这里做了
% As claimed in previous work~\citep{}, knowledge enhancement may be useless or harmful to the performance, we think it may partly because the coverage of knowledge, that is, the more training or testing sample associated with knowledge, the more boost the model can achieve.

\subsection{Limitations}
As show in Eq.~\ref{prediction_layer}, \ourmodel~takes a multi-task loss, these losses introduce parameters, i.e., $W_2, b_2, W_3, b_3$, to linearly transform the representation to the distribution of target classes. Besides, the dimension size and the semantic space of the backbone and knowledge model are different, and the map from the former to the latter introduces parameters in each fusion module, i.e., $W_{(k)} \in \mathbb{R}^{d_3 \times d_2}, 1 \leq k \in \mathbb{N}^+ \leq K$. Moreover, though the parameters of the knowledge module are frozen, it induces computations and time costs in the inference phase, and this problem can be solved by pre-computation~\citep{borgeaud2021improving}. Specifically, we pre-compute each knowledge layer representation of all entity descriptions and cache the pre-computed representations for later use.

\section{Conclusion}
This paper presents a novel architecture ~\ourmodel~for enhancing text representation with entities and entity descriptions. Long description text is represented separately by a lighter knowledge module and then injected to the backbone for knowledge enhancement. On top of the architecture, two entity/description-related auxiliary tasks are introduced to narrow the semantic gap between involved different representations.
Empirical results on knowledge-related and common tasks show the effectiveness of \ourmodel~compared to current state-of-the-art knowledge enhanced methods.
We also conduct extensive ablation studies to demonstrate the impacts of each design choice in ~\ourmodel.
One limitation of our work is that the knowledge module costs computation resources and increases inference time, and it can be easily solved by pre-computation, we leave this for future work.
We believe that \ourmodel~can provide the NLP community with a new way to utilize knowledge for natural language and thus produce better representations.

\section{REFERENCES}

\bibliography{anthology}

\begin{thebibliography}{40}
\expandafter\ifx\csname natexlab\endcsname\relax\def\natexlab#1{#1}\fi

\bibitem[{Bordes et~al.(2013)Bordes, Usunier, Garcia-Duran, Weston, and
  Yakhnenko}]{AntoineBordes2013TranslatingEF}
Antoine Bordes, Nicolas Usunier, Alberto Garcia-Duran, Jason Weston, and Oksana
  Yakhnenko. 2013.
\newblock Translating embeddings for modeling multi-relational data.
\newblock In \emph{Neural Information Processing Systems}.

\bibitem[{Borgeaud et~al.(2021)Borgeaud, Mensch, Hoffmann, Cai, Rutherford,
  Millican, Driessche, Lespiau, Damoc, Clark et~al.}]{borgeaud2021improving}
Sebastian Borgeaud, Arthur Mensch, Jordan Hoffmann, Trevor Cai, Eliza
  Rutherford, Katie Millican, George van~den Driessche, Jean-Baptiste Lespiau,
  Bogdan Damoc, Aidan Clark, et~al. 2021.
\newblock Improving language models by retrieving from trillions of tokens.
\newblock \emph{arXiv preprint arXiv:2112.04426}.

\bibitem[{Br{\"u}mmer et~al.(2016)Br{\"u}mmer, Dojchinovski, and
  Hellmann}]{brummer2016dbpedia}
Martin Br{\"u}mmer, Milan Dojchinovski, and Sebastian Hellmann. 2016.
\newblock Dbpedia abstracts: A large-scale, open, multilingual nlp training
  corpus.
\newblock In \emph{Proceedings of the Tenth International Conference on
  Language Resources and Evaluation (LREC'16)}, pages 3339--3343.

\bibitem[{Choi et~al.(2018)Choi, Levy, Choi, and Zettlemoyer}]{choi2018ultra}
Eunsol Choi, Omer Levy, Yejin Choi, and Luke Zettlemoyer. 2018.
\newblock Ultra-fine entity typing.
\newblock In \emph{Proceedings of the 56th Annual Meeting of the Association
  for Computational Linguistics (Volume 1: Long Papers)}, pages 87--96.

\bibitem[{Chuang et~al.(2022)Chuang, Dangovski, Luo, Zhang, Chang,
  Solja{\v{c}}i{\'c}, Li, Yih, Kim, and Glass}]{chuang2022diffcse}
Yung-Sung Chuang, Rumen Dangovski, Hongyin Luo, Yang Zhang, Shiyu Chang, Marin
  Solja{\v{c}}i{\'c}, Shang-Wen Li, Wen-tau Yih, Yoon Kim, and James Glass.
  2022.
\newblock Diffcse: Difference-based contrastive learning for sentence
  embeddings.
\newblock \emph{arXiv preprint arXiv:2204.10298}.

\bibitem[{Devlin et~al.(2019)Devlin, Chang, Lee, and
  Toutanova}]{devlin2019bert}
Jacob Devlin, Ming-Wei Chang, Kenton Lee, and Kristina Toutanova. 2019.
\newblock Bert: Pre-training of deep bidirectional transformers for language
  understanding.
\newblock In \emph{Proceedings of the 2019 Conference of the North American
  Chapter of the Association for Computational Linguistics: Human Language
  Technologies, Volume 1 (Long and Short Papers)}, pages 4171--4186.

\bibitem[{Ferragina and Scaiella(2010)}]{ferragina2010tagme}
Paolo Ferragina and Ugo Scaiella. 2010.
\newblock Tagme: on-the-fly annotation of short text fragments (by wikipedia
  entities).
\newblock In \emph{Proceedings of the 19th ACM international conference on
  Information and knowledge management}, pages 1625--1628.

\bibitem[{Gao et~al.(2018)Gao, Liang, Han, Yakout, and
  Mohamed}]{gao2018building}
Yuqing Gao, Jisheng Liang, Benjamin Han, Mohamed Yakout, and Ahmed Mohamed.
  2018.
\newblock Building a large-scale, accurate and fresh knowledge graph.
\newblock \emph{KDD-2018, Tutorial}, 39:1939--1374.

\bibitem[{Han et~al.(2018)Han, Zhu, Yu, Wang, Yao, Liu, and
  Sun}]{han2018fewrel}
Xu~Han, Hao Zhu, Pengfei Yu, Ziyun Wang, Yuan Yao, Zhiyuan Liu, and Maosong
  Sun. 2018.
\newblock Fewrel: A large-scale supervised few-shot relation classification
  dataset with state-of-the-art evaluation.
\newblock In \emph{Proceedings of the 2018 Conference on Empirical Methods in
  Natural Language Processing}, pages 4803--4809.

\bibitem[{Levine et~al.(2020)Levine, Lenz, Dagan, Ram, Padnos, Sharir,
  Shalev-Shwartz, Shashua, and Shoham}]{levine2020sensebert}
Yoav Levine, Barak Lenz, Or~Dagan, Ori Ram, Dan Padnos, Or~Sharir, Shai
  Shalev-Shwartz, Amnon Shashua, and Yoav Shoham. 2020.
\newblock Sensebert: Driving some sense into bert.
\newblock In \emph{Proceedings of the 58th Annual Meeting of the Association
  for Computational Linguistics}, pages 4656--4667.

\bibitem[{Li{\'e}tard et~al.(2021)Li{\'e}tard, Abdou, and
  S{\o}gaard}]{lietard2021language}
Bastien Li{\'e}tard, Mostafa Abdou, and Anders S{\o}gaard. 2021.
\newblock Do language models know the way to rome?
\newblock In \emph{Proceedings of the Fourth BlackboxNLP Workshop on Analyzing
  and Interpreting Neural Networks for NLP}, pages 510--517.

\bibitem[{Ling et~al.(2015)Ling, Singh, and Weld}]{ling2015design}
Xiao Ling, Sameer Singh, and Daniel~S Weld. 2015.
\newblock Design challenges for entity linking.
\newblock \emph{Transactions of the Association for Computational Linguistics},
  3:315--328.

\bibitem[{Liu et~al.(2019{\natexlab{a}})Liu, Gardner, Belinkov, Peters, and
  Smith}]{NelsonFLiu2019LinguisticKA}
Nelson~F. Liu, Matt Gardner, Yonatan Belinkov, Matthew~E. Peters, and Noah~A.
  Smith. 2019{\natexlab{a}}.
\newblock Linguistic knowledge and transferability of contextual
  representations.
\newblock In \emph{North American Chapter of the Association for Computational
  Linguistics}.

\bibitem[{Liu et~al.(2019{\natexlab{b}})Liu, Ott, Goyal, Du, Joshi, Chen, Levy,
  Lewis, Zettlemoyer, and Stoyanov}]{liu2019roberta}
Yinhan Liu, Myle Ott, Naman Goyal, Jingfei Du, Mandar Joshi, Danqi Chen, Omer
  Levy, Mike Lewis, Luke Zettlemoyer, and Veselin Stoyanov. 2019{\natexlab{b}}.
\newblock Roberta: A robustly optimized bert pretraining approach.
\newblock \emph{arXiv preprint arXiv:1907.11692}.

\bibitem[{Loshchilov and Hutter(2018)}]{loshchilov2018decoupled}
Ilya Loshchilov and Frank Hutter. 2018.
\newblock Decoupled weight decay regularization.
\newblock In \emph{International Conference on Learning Representations}.

\bibitem[{Micikevicius et~al.(2018)Micikevicius, Narang, Alben, Diamos, Elsen,
  Garcia, Ginsburg, Houston, Kuchaiev, Venkatesh
  et~al.}]{micikevicius2018mixed}
Paulius Micikevicius, Sharan Narang, Jonah Alben, Gregory Diamos, Erich Elsen,
  David Garcia, Boris Ginsburg, Michael Houston, Oleksii Kuchaiev, Ganesh
  Venkatesh, et~al. 2018.
\newblock Mixed precision training.
\newblock In \emph{International Conference on Learning Representations}.

\bibitem[{Miller(1995)}]{miller1995wordnet}
George~A Miller. 1995.
\newblock Wordnet: a lexical database for english.
\newblock \emph{Communications of the ACM}, 38(11):39--41.

\bibitem[{Orr et~al.(2020{\natexlab{a}})Orr, Leszczynski, Arora, Wu, Guha,
  Ling, and R{\'e}}]{LaurelOrr2020BootlegCT}
Laurel Orr, Megan Leszczynski, Simran Arora, Sen Wu, Neel Guha, Xiao Ling, and
  Christopher R{\'e}. 2020{\natexlab{a}}.
\newblock Bootleg: Chasing the tail with self-supervised named entity
  disambiguation.
\newblock In \emph{Conference on Innovative Data Systems Research}.

\bibitem[{Orr et~al.(2020{\natexlab{b}})Orr, Leszczynski, Arora, Wu, Guha,
  Ling, and Re}]{orr2020bootleg}
Laurel Orr, Megan Leszczynski, Simran Arora, Sen Wu, Neel Guha, Xiao Ling, and
  Christopher Re. 2020{\natexlab{b}}.
\newblock Bootleg: chasing the tail with self-supervised named entity
  disambiguation.
\newblock \emph{arXiv preprint arXiv:2010.10363}.

\bibitem[{Peters et~al.(2019)Peters, Neumann, Logan, Schwartz, Joshi, Singh,
  and Smith}]{peters2019knowledge}
Matthew~E Peters, Mark Neumann, Robert Logan, Roy Schwartz, Vidur Joshi, Sameer
  Singh, and Noah~A Smith. 2019.
\newblock Knowledge enhanced contextual word representations.
\newblock In \emph{Proceedings of the 2019 Conference on Empirical Methods in
  Natural Language Processing and the 9th International Joint Conference on
  Natural Language Processing (EMNLP-IJCNLP)}, pages 43--54.

\bibitem[{Petroni et~al.(2019)Petroni, Rockt{\"a}schel, Riedel, Lewis, Bakhtin,
  Wu, and Miller}]{petroni2019language}
Fabio Petroni, Tim Rockt{\"a}schel, Sebastian Riedel, Patrick Lewis, Anton
  Bakhtin, Yuxiang Wu, and Alexander Miller. 2019.
\newblock Language models as knowledge bases?
\newblock In \emph{Proceedings of the 2019 Conference on Empirical Methods in
  Natural Language Processing and the 9th International Joint Conference on
  Natural Language Processing (EMNLP-IJCNLP)}, pages 2463--2473.

\bibitem[{Rajpurkar et~al.(2016)Rajpurkar, Zhang, Lopyrev, and
  Liang}]{PranavRajpurkar2016SQuAD1Q}
Pranav Rajpurkar, Jian Zhang, Konstantin Lopyrev, and Percy Liang. 2016.
\newblock Squad: 100,000+ questions for machine comprehension of text.
\newblock \emph{empirical methods in natural language processing}.

\bibitem[{Sanh et~al.(2019)Sanh, Debut, Chaumond, and
  Wolf}]{sanh2019distilbert}
Victor Sanh, Lysandre Debut, Julien Chaumond, and Thomas Wolf. 2019.
\newblock Distilbert, a distilled version of bert: smaller, faster, cheaper and
  lighter.
\newblock \emph{arXiv preprint arXiv:1910.01108}.

\bibitem[{Socher et~al.(2013)Socher, Perelygin, Wu, Chuang, Manning, Ng, and
  Potts}]{RichardSocher2013RecursiveDM}
Richard Socher, Alex Perelygin, Jean~Y. Wu, Jason Chuang, Christopher~D.
  Manning, Andrew~Y. Ng, and Christopher Potts. 2013.
\newblock Recursive deep models for semantic compositionality over a sentiment
  treebank.
\newblock \emph{empirical methods in natural language processing}.

\bibitem[{Speer et~al.(2017)Speer, Chin, and Havasi}]{speer2017conceptnet}
Robyn Speer, Joshua Chin, and Catherine Havasi. 2017.
\newblock Conceptnet 5.5: An open multilingual graph of general knowledge.
\newblock In \emph{Thirty-first AAAI conference on artificial intelligence}.

\bibitem[{van Hulst et~al.(2020)van Hulst, Hasibi, Dercksen, Balog, and
  de~Vries}]{van2020rel}
Johannes~M van Hulst, Faegheh Hasibi, Koen Dercksen, Krisztian Balog, and
  Arjen~P de~Vries. 2020.
\newblock Rel: An entity linker standing on the shoulders of giants.
\newblock In \emph{Proceedings of the 43rd International ACM SIGIR Conference
  on Research and Development in Information Retrieval}, pages 2197--2200.

\bibitem[{Vaswani et~al.(2017)Vaswani, Shazeer, Parmar, Uszkoreit, Jones,
  Gomez, Kaiser, and Polosukhin}]{vaswani2017attention}
Ashish Vaswani, Noam Shazeer, Niki Parmar, Jakob Uszkoreit, Llion Jones,
  Aidan~N Gomez, {\L}ukasz Kaiser, and Illia Polosukhin. 2017.
\newblock Attention is all you need.
\newblock \emph{Advances in neural information processing systems}, 30.

\bibitem[{Vrande{\v{c}}i{\'c} and Kr{\"o}tzsch(2014)}]{vrandevcic2014wikidata}
Denny Vrande{\v{c}}i{\'c} and Markus Kr{\"o}tzsch. 2014.
\newblock Wikidata: a free collaborative knowledgebase.
\newblock \emph{Communications of the ACM}, 57(10):78--85.

\bibitem[{Wang et~al.(2018)Wang, Singh, Michael, Hill, Levy, and
  Bowman}]{wang2018glue}
Alex Wang, Amanpreet Singh, Julian Michael, Felix Hill, Omer Levy, and Samuel
  Bowman. 2018.
\newblock Glue: A multi-task benchmark and analysis platform for natural
  language understanding.
\newblock In \emph{Proceedings of the 2018 EMNLP Workshop BlackboxNLP:
  Analyzing and Interpreting Neural Networks for NLP}, pages 353--355.

\bibitem[{Wang et~al.(2021{\natexlab{a}})Wang, Tang, Duan, Wei, Huang, Ji, Cao,
  Jiang, and Zhou}]{wang2021k}
Ruize Wang, Duyu Tang, Nan Duan, Zhongyu Wei, Xuan-Jing Huang, Jianshu Ji,
  Guihong Cao, Daxin Jiang, and Ming Zhou. 2021{\natexlab{a}}.
\newblock K-adapter: Infusing knowledge into pre-trained models with adapters.
\newblock In \emph{Findings of the Association for Computational Linguistics:
  ACL-IJCNLP 2021}, pages 1405--1418.

\bibitem[{Wang et~al.(2021{\natexlab{b}})Wang, Gao, Zhu, Zhang, Liu, Li, and
  Tang}]{wang2021kepler}
Xiaozhi Wang, Tianyu Gao, Zhaocheng Zhu, Zhengyan Zhang, Zhiyuan Liu, Juanzi
  Li, and Jian Tang. 2021{\natexlab{b}}.
\newblock Kepler: A unified model for knowledge embedding and pre-trained
  language representation.
\newblock \emph{Transactions of the Association for Computational Linguistics},
  9:176--194.

\bibitem[{Wu et~al.(2020)Wu, Petroni, Josifoski, Riedel, and
  Zettlemoyer}]{wu2020scalable}
Ledell Wu, Fabio Petroni, Martin Josifoski, Sebastian Riedel, and Luke
  Zettlemoyer. 2020.
\newblock Scalable zero-shot entity linking with dense entity retrieval.
\newblock In \emph{Proceedings of the 2020 Conference on Empirical Methods in
  Natural Language Processing (EMNLP)}, pages 6397--6407.

\bibitem[{Xiong et~al.(2019)Xiong, Du, Wang, and
  Stoyanov}]{xiong2019pretrained}
Wenhan Xiong, Jingfei Du, William~Yang Wang, and Veselin Stoyanov. 2019.
\newblock Pretrained encyclopedia: Weakly supervised knowledge-pretrained
  language model.
\newblock In \emph{International Conference on Learning Representations}.

\bibitem[{Xu et~al.(2021{\natexlab{a}})Xu, Fang, Zhu, and Zeng}]{xu2021does}
Ruochen Xu, Yuwei Fang, Chenguang Zhu, and Michael Zeng. 2021{\natexlab{a}}.
\newblock Does knowledge help general nlu? an empirical study.
\newblock \emph{arXiv preprint arXiv:2109.00563}.

\bibitem[{Xu et~al.(2021{\natexlab{b}})Xu, Zhu, Xu, Liu, Zeng, and
  Huang}]{xu2021fusing}
Yichong Xu, Chenguang Zhu, Ruochen Xu, Yang Liu, Michael Zeng, and Xuedong
  Huang. 2021{\natexlab{b}}.
\newblock Fusing context into knowledge graph for commonsense question
  answering.
\newblock In \emph{Findings of the Association for Computational Linguistics:
  ACL-IJCNLP 2021}, pages 1201--1207.

\bibitem[{Yamada et~al.(2020)Yamada, Asai, Shindo, Takeda, and
  Matsumoto}]{yamada2020luke}
Ikuya Yamada, Akari Asai, Hiroyuki Shindo, Hideaki Takeda, and Yuji Matsumoto.
  2020.
\newblock Luke: Deep contextualized entity representations with entity-aware
  self-attention.
\newblock In \emph{Proceedings of the 2020 Conference on Empirical Methods in
  Natural Language Processing (EMNLP)}, pages 6442--6454.

\bibitem[{Yang et~al.(2021)Yang, Liu, Xiao, Li, Lian, Agrawal, Singh, Sun, and
  Xie}]{yang2021graphformers}
Junhan Yang, Zheng Liu, Shitao Xiao, Chaozhuo Li, Defu Lian, Sanjay Agrawal,
  Amit Singh, Guangzhong Sun, and Xing Xie. 2021.
\newblock Graphformers: Gnn-nested transformers for representation learning on
  textual graph.
\newblock \emph{Advances in Neural Information Processing Systems}, 34.

\bibitem[{Zhang et~al.(2017)Zhang, Zhong, Chen, Angeli, and
  Manning}]{zhang2017position}
Yuhao Zhang, Victor Zhong, Danqi Chen, Gabor Angeli, and Christopher~D Manning.
  2017.
\newblock Position-aware attention and supervised data improve slot filling.
\newblock In \emph{Proceedings of the 2017 Conference on Empirical Methods in
  Natural Language Processing}, pages 35--45.

\bibitem[{Zhang et~al.(2019)Zhang, Han, Liu, Jiang, Sun, and
  Liu}]{zhang2019ernie}
Zhengyan Zhang, Xu~Han, Zhiyuan Liu, Xin Jiang, Maosong Sun, and Qun Liu. 2019.
\newblock Ernie: Enhanced language representation with informative entities.
\newblock In \emph{Proceedings of the 57th Annual Meeting of the Association
  for Computational Linguistics}, pages 1441--1451.

\bibitem[{Zhao et~al.(2022)Zhao, Ma, and Ren}]{zhao2022kesa}
Qinghua Zhao, Shuai Ma, and Shuo Ren. 2022.
\newblock Kesa: A knowledge enhanced approach for sentiment analysis.
\newblock \emph{arXiv preprint arXiv:2202.12093}.

\end{thebibliography}
\bibliographystyle{acl_natbib}

\appendix

\section{Appendix}
\label{sec:appendix}
\subsection{Implementation Details}\label{implementation_details}

% %%%%%%%%%%%%%%%%%%%%%%%%%%%%%%%%%%%%%%%%%%%%%%%%%%%%
\begin{table*}%[t]
\centering
\resizebox{.95\textwidth}{!}{
\begin{tabular}{l|cccccc}
\hline Name&   \openentity& \figer&    \fewrel&    \tacred& \sst&   \eem \\ 
Batch size& 4&  2048&   32& 32& 128&    128\\
Maximum text length& 256&   128&    256&    256&    128&    32\\
Maximum description length& 64& 64& 64& 64& 64& 32\\
Learning rate&  1e-5&   2e-5&   1e-5&   1e-5&    \multicolumn{2}{c}{\{1e-5, 2e-5, 3e-5, 5e-5\}}\\
Epoch&  3&  2&  10& 5&  3&  3\\
Evaluation steps&   per epoch&  50& per epoch&  500&    500&    500\\
Warmup ratio&   0.06&   0.06&   0.1&    0.1&    0.1&    0.1\\
Number of entities&     4&  2&  4&  4&  4&  2\\
Number of descriptions& 1&  1&  1&  1&  1&  1\\
$\alpha$&   1.0&    1.0&    1.0&    1.0&    1.0&    1.0\\
$\beta$&    0.01&   0.01&   1.0&   0.1&   0.1&   1.0\\
Alignment&  \multicolumn{6}{c}{the top six backbone layers are connected to knowledge layers}\\
Recognized entities&    \multicolumn{4}{c}{ERNIE-THU}&    TAGME&  Satori\\
Times of experiments&   20& 1&  20& 5&  4&  4 \\
Reported results&   average&    -&  average&    average&    best&   best    \\
\hline
\end{tabular}
}
\caption{Hyper-parameters and other details of our experiments. ``average'' and ``best'' refer to that the averaged/best results are reported.}
\label{hyper_parameters} 
\end{table*}
% %%%%%%%%%%%%%%%%%%%%%%%%%%%%%%%%%%%%%%%%%%%%%%%%%%%%

For most parameters, we adopt the value recommended in LUKE, and the parameters we use are listed in Table~\ref{hyper_parameters}.
We optimized the model by AdamW~\cite{loshchilov2018decoupled}, and a linear learning rate decay is adopted.
Besides, mixed precision~\citep{micikevicius2018mixed} is adopted to accelerate computation.
The number of associated entities and descriptions is searched from \{1, 2, 4, 6, 8\}, and the $\alpha$ and $\beta$ are searched from \{1.0, 0.1, 0.01\}.
In our experiments, four entities, one description and $\alpha=1.0$ are used as default. And the knowledge layer is aligned to the last $K$ backbone layers.
Since entities are used, we need entity linker~\citep{wu2020scalable, orr2020bootleg, van2020rel, ferragina2010tagme} to recognize the entities included in the text. In our experiment, we adopt the linked datasets provided by~\citep{zhang2019ernie}, and for ~\sst, TAGME~\cite{ferragina2010tagme} is used to perform entity linking. For \eem, entity and entity description are given, which is extracted from Microsoft knowledge graph Satori~\citep{gao2018building}.
Positive entities are recognized by the entity linker, while negative entities are randomly sampled from entity vocabulary. When no entity is included in one sentence, an entity identifier ``[MASK]'' is used as a positive entity.
For \eem~and \figer~dataset, considering its large training samples, we run it just for one time. 
Note that, considering the large-scale training set of \figer~dataset, to accelerate the training process, we reduce the maximum sequence length from 256 to 128. Besides, with 2 million training samples and only 500 test samples, it is easy to overfit, and we used the parameters recommended in ~\citep{zhang2019ernie, wang2021k} and did not do a grid-search. Specifically, the batch size per GPU is 64, the gradient accumulation step is set to 8, four NVIDIA V100 of 32G are used, and then it takes about two hours per epoch and the best results are always obtained in step 300, the learning rate is set to 2e-5, the warmup step is set to 6\%, $\alpha=1.0, \beta=0.01$, the number of entities and descriptions are set to 2 and 1, respectively. We run training on \figer~for two epochs and evaluate it every 50 steps.

% %%%%%%%%%%%%%%%%%%%%%%%%%%%%%%%%%%%%%%%%%%%%%%%%%%%%
% \begin{table}%[t]
% \centering
% % \resizebox{.95\columnwidth}{!}{
% \begin{tabular}{l|l}
% \hline Name&   XXX \\ 

% kt-attn&    \\
% roberta-base&   \\
% kepler& \\
% \hline
% roberta-large&  \\
% kadapter&   \\
% luke&   \\
% \ourmodel~(w/o K)&  \\
% \ourmodel&  \\
% \ourmodel~(R)&  \\

% \hline
% \end{tabular}
% % }
% \caption{Results of aspect-based sentiment analysis on the \absa~dataset.}
% \label{results_absa} 
% \end{table}
% %%%%%%%%%%%%%%%%%%%%%%%%%%%%%%%%%%%%%%%%%%%%%%%%%%%%

\end{document}